\documentclass[runningheads]{llncs}
\usepackage{graphicx}
\usepackage{amsmath,amssymb} 
\usepackage{color}

\usepackage{makecell}
\usepackage{wrapfig}
\usepackage{hyperref}
\usepackage{soul,xcolor}

\begin{document}
\title{Video Object Detection with an Aligned Spatial-Temporal Memory} 

\titlerunning{Video Object Detection with an Aligned Spatial-Temporal Memory}
%
\author{Fanyi Xiao\orcidID{0000-0002-9839-1139} \and
Yong Jae Lee\orcidID{0000-0001-9863-1270}}
%
\authorrunning{F. Xiao and Y. J. Lee}
%

\institute{University of California, Davis\\
\email{\{fyxiao,yongjaelee\}@ucdavis.edu}}
\maketitle              
\begin{abstract}
We introduce Spatial-Temporal Memory Networks for video object detection. At its core, a novel Spatial-Temporal Memory module (STMM) serves as the recurrent computation unit to model long-term temporal appearance and motion dynamics.  The STMM's design enables full integration of pretrained backbone CNN weights, which we find to be critical for accurate detection. Furthermore, in order to tackle object motion in videos, we propose a novel MatchTrans module to align the spatial-temporal memory from frame to frame.  Our method produces state-of-the-art results on the benchmark ImageNet VID dataset, and our ablative studies clearly demonstrate the contribution of our different design choices. We release our code and models at \textcolor{blue}{\url{http://fanyix.cs.ucdavis.edu/project/stmn/project.html}}.

\end{abstract}

\section{Introduction}

Object detection is a fundamental problem in computer vision.  While there has been a long history of detecting objects in \emph{static images}, there has been much less research in detecting objects in \emph{videos}.  However, cameras on robots, surveillance systems, vehicles, wearable devices, etc., receive videos instead of static images.  Thus, for these systems to recognize the key objects and their interactions, it is critical that they be equipped with accurate \emph{video} object detectors.

\begin{figure*}[t]
	\centering
	\includegraphics[width=1\textwidth]{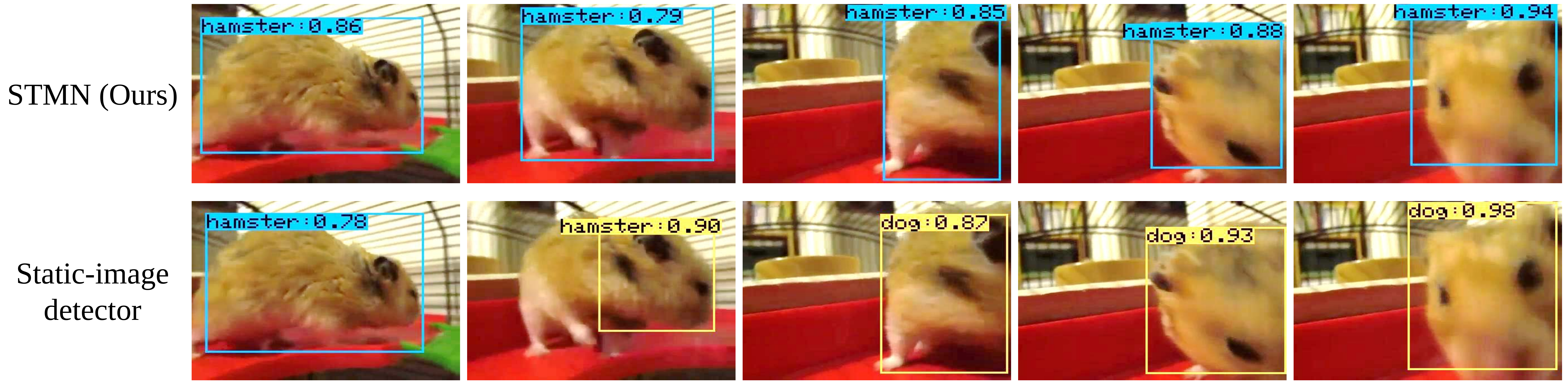}
	\caption{Static image detectors (such as Fast-RCNN~\cite{fast-rcnn} or R-FCN~\cite{dai-nips2016}), tend to fail under occlusion or extreme pose (false detections shown in yellow). By learning to aggregate information across time, our STMN video object detector can produce correct detections in frames with challenging pose/viewpoints.  In this example, it aggregates information from the easier profile views of the hamster (first two frames) to aid detection in occluded or extreme views of the hamster (third-fifth frames).}
	\label{fig:concept}
\end{figure*}

The simplest way to detect objects in video is to run a static image-based detector independently on each frame.  However, due to the different biases and challenges of video (e.g., motion blur, low-resolution, compression artifacts), an image detector usually does not generalize well.  More importantly, videos provide \emph{rich temporal and motion information} that should be utilized by the detector during both training and testing.  For example, in Fig.~\ref{fig:concept}, since the hamster's profile view (frames 1-2) is much easier to detect than the challenging viewpoint/pose in later frames, the image detector only succeeds in detecting the leading frame of the sequence. On the other hand, by learning to aggregate useful information over time, a video object detector can robustly detect the object under extreme viewpoint/pose.  

Therefore, in recent years, there has been a growing interest in the community on designing video object detectors~\cite{han-imagenetvid2015,han-arxiv2016,lee-eccv2016,kang-tcnn2017,feichtenhofer-iccv2017,zhu-iccv2017,zhu-cvpr2018}. However, many existing methods exploit temporal information in an ad-hoc, post-processing manner -- static object detections returned by an image detector like R-FCN~\cite{dai-nips2016} or Faster R-CNN~\cite{ren-nips2015} are linked across frames~\cite{han-arxiv2016,lee-eccv2016,kang-tcnn2017}, or video segmentation is performed to refine the detection results~\cite{han-imagenetvid2015}.   Although these methods show improvement over a static image detector, exploiting temporal information as post-processing is sub-optimal since temporal and motion information are ignored during detector training. As such, they have difficulty overcoming consecutive failures of the static detector e.g., when the object-of-interest has large occlusion or unusual appearance for a long time.

More recent works~\cite{feichtenhofer-iccv2017,zhu-iccv2017,zhu-cvpr2018} learn to exploit temporal information \emph{during training} by either learning to combine features across neighboring frames or by predicting the displacement of detection boxes across frames.  However, these methods operate on fixed-length temporal windows and thus have difficulty modeling \emph{variable and long-term} temporal information. While the Tubelet Proposal Network~\cite{kang-cvpr2017} does model long-term dependencies, it uses \emph{vectors} to represent the memory of the recurrent unit, and hence loses spatial information.  To compensate, it computes the memory vectors at the region-level for each tube (sequence of proposals), but this can be very slow and depends strongly on having accurate initial tubes.

To address these limitations, we introduce the \emph{Spatial-Temporal Memory Network} (STMN), which jointly learns to model and align an object's long-term appearance and motion dynamics in an end-to-end fashion for video object detection.  At its core is the Spatial-Temporal Memory Module (STMM), which is a convolutional recurrent computation unit that fully integrates pre-trained weights learned from static images (e.g., ImageNet~\cite{imagenet}).  This design choice is critical in addressing the practical challenge of learning from contemporary video datasets, which largely lack intra-category object diversity; i.e., since video frames are highly redundant, a video dataset of e.g., 1 million frames has much lower diversity than an image dataset with 1 million images.  By designing our memory unit to be compatible with pre-trained weights from both its preceding and succeeding layers, we show that it outperforms the standard ConvGRU~\cite{ballas-iclr2016} recurrent module for video object detection.

Furthermore, in order to account for the 2D spatial nature of visual data, the STMM preserves the spatial information of each frame in its memory.  In particular, to achieve accurate pixel-level spatial alignment over time, the STMM uses a novel MatchTrans module to explicitly model the displacement introduced by motion across frames.  Since the convolutional features for each frame are aligned and aggregated in the spatial memory, the feature for any particular object region is well-localized and contains information across multiple frames.  Furthermore, each region feature can be extracted trivially via ROI pooling from the memory.

In summary, our main contribution is a novel spatial-temporal memory network for video object detection.  Our ablative studies show the benefits provided by the STMM and MatchTrans modules -- integrating pre-trained static image weights and providing spatial alignment across time.  These design choices lead to state-of-the-art results on the ImageNet video object detection dataset (VID)~\cite{imagenetvid} across different base detectors and backbone networks.

\section{Related work}

\paragraph{Static image object detection.}  Recent work that adopt deep neural networks have significantly advanced the state-of-the-art in static image object detection~\cite{rcnn,overfeat,fast-rcnn,ren-nips2015,redmon-yolo2016,liu-ssd2016,dai-nips2016,shrivastava-eccv2016,chen-iccv2017}.  Our work also builds on the success of deep networks to learn the features, classifier, and bounding box localizer in an end-to-end framework. However, in contrast to most existing work that focus on detecting objects in static images, this paper aims to detect objects in \emph{videos}.

\paragraph{Video object detection.} Compared to static image-based object detection, there has been less research in detecting objects in videos.  Early work processed videos captured from a static camera or made strong assumptions about the type of scene (e.g., highway traffic camera for detecting cars or an indoor room for detecting persons)~\cite{pfinder,coifman-1998}.  Later work used hand-designed features by aggregating simple motion cues (based on optical flow, temporal differences, or tracking), and focused mostly on pedestrian detection~\cite{viola-ijcv2005,dalal-eccv2006,jones-icpr2008,park-cvpr2013}.

With the introduction of ImageNet VID~\cite{imagenetvid} in 2015, researchers have focused on more generic categories and realistic videos.  However, many existing approaches combine per-frame detections from a static image detector via tracking in a two-stage pipeline~\cite{han-arxiv2016,tripathi-bmvc2016,kang-tcnn2017}. Since the motion and temporal cues are used as a post-processing step only during testing, many heuristic choices are required, which can lead to sub-optimal results. In contrast, our approach directly \emph{learns} to integrate the motion and temporal dependencies during training.  Our end-to-end architecture also leads to a clean and fast runtime.

Sharing our goal of leveraging temporal information \emph{during training}, the recent works of Zhu et al.~\cite{zhu-iccv2017,zhu-cvpr2018} learn to combine features of different frames with a feed-forward network for improved detection accuracy.  Our method differs in that it produces a \emph{spatial-temporal memory} that can carry on information across long and variable number of frames, whereas the methods in~\cite{zhu-iccv2017,zhu-cvpr2018} can only aggregate information over a small and fixed number of frames. In Sec.~\ref{sect:windowlength}, we demonstrate the benefits gained from this flexibility.  Although the approach of Kang et al.~\cite{kang-cvpr2017} uses memory to aggregate temporal information, it uses a vector representation.  Since spatial information is lost, it computes a separate memory vector for each region tube (sequence of proposals) which can make the approach very slow.  In contrast, our approach only needs to compute a single \emph{frame-level} spatial memory, whose computation is independent of the number of proposals.

Finally, Detect and Track~\cite{feichtenhofer-iccv2017} aims to unify detection and tracking, where the correlation between consecutive {\it two} frames are used to predict the movement of the detection boxes. Unlike~\cite{feichtenhofer-iccv2017}, our spatial-temporal memory aggregates information across $t>2$ frames. Furthermore, while our approach also computes the correlation between neighboring frames with the proposed MatchTrans module, we use it to warp the entire feature map for alignment (i.e., at the coarse pixel-level), rather than use it to predict the displacement of the boxes.  Overall, these choices lead to state-of-the-art detection accuracy on ImageNet VID.

\paragraph{Learning with videos.} Apart from video object detection, other recent work use convolutional and/or recurrent networks for video classification~\cite{karpathy-cvpr2014,tran-iccv2015,ballas-iclr2016}.  These methods tend to model entire video frames instead of pixels, which means the fine-grained details required for localizing objects are often lost. Object tracking (e.g.,~\cite{li-cvpr2015,nam-cvpr2016}), which requires accurate localization, is also closely-related. The key difference is that in tracking, the bounding box of the first frame is given and the tracker does not necessarily need to know the semantic category of the object being tracked.

\paragraph{Modeling sequence data with RNNs.} In computer vision, RNNs have been used for image captioning~\cite{kiros-arxiv2014,vinyals-cvpr2015,donahue-cvpr2015}, visual attention~\cite{ba-attention2014,mnih-nips2014,xu-arxiv2015}, action/object recognition~\cite{donahue-cvpr2015,ballas-iclr2016}, human pose estimation~\cite{fragkiadaki-iccv2015,carreira-cvpr2016}, and semantic segmentation~\cite{zheng-iccv2015}. Recently, Tripathi et al.~\cite{tripathi-bmvc2016} adopted RNNs for video object detection. However, in their pipeline, the CNN-based detector is first trained, then an RNN is trained to refine the detection outputs of the CNN.

Despite the wide adoption of RNNs in various vision tasks, most approaches work with vector-form memory units (as in standard LSTM/GRU). To take spatial locality into account, Ballas et al.~\cite{ballas-iclr2016} proposed convolutional gated recurrent units (ConvGRU) and applied it to the task of action recognition. Built upon~\cite{ballas-iclr2016}, Tokmakov et al.~\cite{tokmakov-iccv2017} used ConvGRUs for the task of video object segmentation. Our work differs in three ways: (1) we classify bounding boxes rather than frames or pixels; (2) we propose a new recurrent computation unit called STMM that makes better use of static image detector weights pre-trained on a large-scale image dataset like ImageNet; and (3) our spatial-temporal memory is aligned frame-to-frame through our MatchTrans module.  We show that these properties lead to better results than ConvGRU for video object detection.
\section{Approach}\label{sect:approach}

We propose a novel RNN architecture called the Spatial-Temporal Memory Network (STMN) to model an object's changing appearance and motion over time for video object detection.

\begin{figure*}[t!]
	\centering
	\includegraphics[width=1\textwidth]{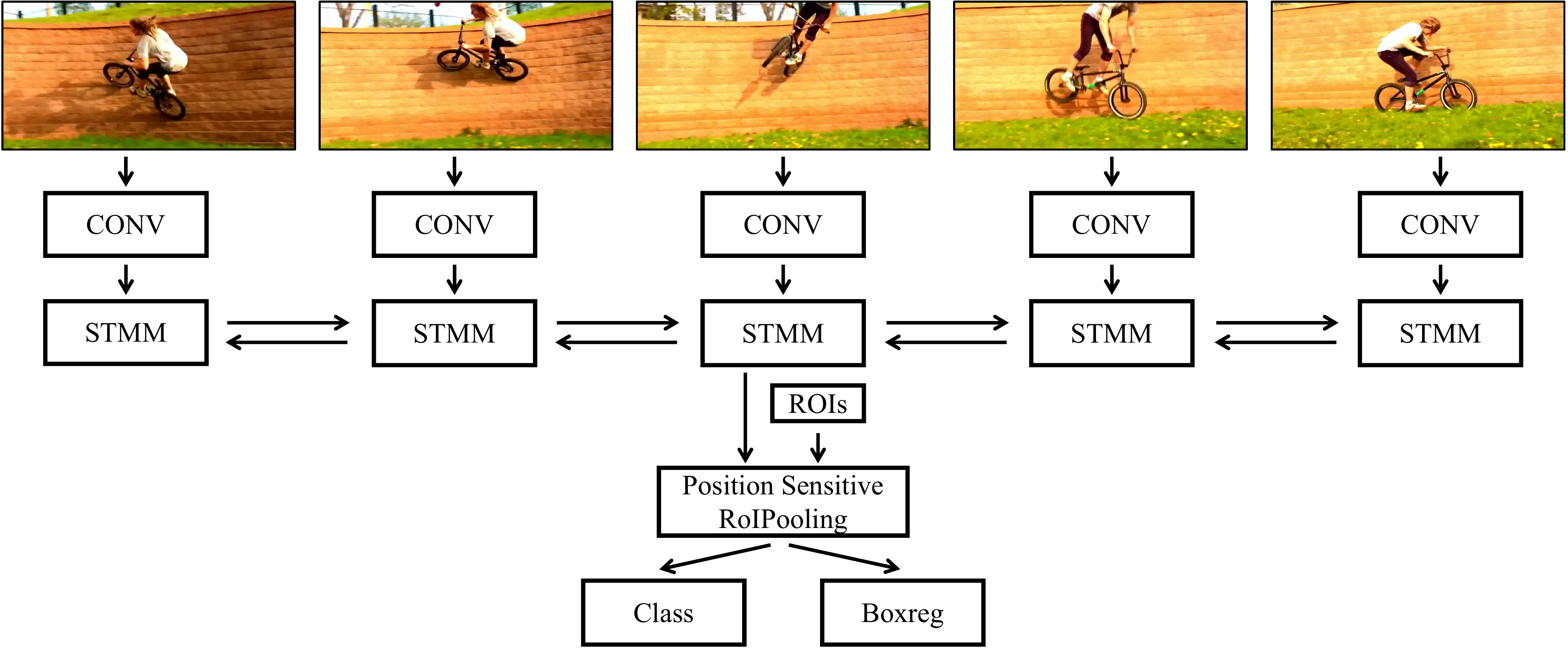}
	\caption{Our STMN architecture. Consecutive frames are forwarded through the convolutional stacks to obtain spatial-preserving convolutional feature maps, which are then fed into the spatial-temporal memory module (STMM). In this example, in order to detect an object on the center frame, information flows into the center STMM from all five frames.   The STMM output from the center frame is then fed into a classification and box regression sub-network. }
	\label{fig:architecture}
\end{figure*}

\subsection{Overview}  The overall architecture is shown in Fig.~\ref{fig:architecture}.  Assuming a video sequence of length $T$, each frame is first forwarded through a convnet to obtain convolutional feature maps $F_1, F_2, ..., F_T$ as appearance features. To aggregate information along the temporal axis, the appearance feature of each frame is fed into the Spatial-Temporal Memory Module (STMM).  The STMM at time step $t$ receives the appearance feature for the current frame $F_t$, as well as a spatial-temporal memory $M^{\rightarrow}_{t-1}$, which carries the information of all previous frames up through timestep $t-1$. The STMM then updates the spatial-temporal memory for the current time step $M^{\rightarrow}_t$ conditioned on both $F_t$ and $M^{\rightarrow}_{t-1}$.  In order to capture information from both previous and later frames, we use two STMMs, one for each direction, to obtain both $M^{\rightarrow}$ and $M^{\leftarrow}$. These are then concatenated to produce the temporally modulated memory $M$ for each frame.

The concatenated memory $M$, which also preserves spatial information, is then fed into subsequent convolution/fully-connected layers for both category classification and bounding box regression.  This way, our approach combines information from both the current frame as well as temporally-neighboring frames when making its detections. This helps, for instance, in the case of detecting a frontal-view bicycle in the center frame of Fig.~\ref{fig:architecture} (which is hard), if we have seen its side-view (which is easier) from nearby frames.  In contrast, a static image detector would only see the frontal-view bicycle when making its detection.

Finally, to train the detector, we use the same loss function used in R-FCN~\cite{dai-nips2016}. Specifically, for each frame in a training sequence, we enforce a cross-entropy loss between the predicted class label and the ground-truth label, and enforce a smooth $L1$ loss on the predicted bounding box regression coefficients. During testing, we slide the testing window and detect on all frames within each sliding window, to be consistent with our training procedure.

\subsection{Spatial-temporal memory module}\label{sect:stmm} We next explain how the STMM models the temporal correlation of an object across frames. At each time step, the STMM takes as input $F_t$ and $M_{t-1}$ and computes the following:
\begin{align}
z_t &= {\tt BN^*}({\tt ReLU}(W_z * F_t + U_z * M_{t-1})), \label{eq:SMMz}\\
r_t &={\tt BN^*}({\tt ReLU}(W_r * F_t + U_r * M_{t-1})), \label{eq:SMMr}\\
\tilde{M}_t &= {\tt ReLU}(W * F_t + U * (M_{t-1} \odot r_t)), \label{eq:SMMCand}\\
M_t &= (1-z_t) \odot M_{t-1} + z_t \odot \tilde{M}_t \label{eq:SMMFinalM},
\end{align}
where $\odot$ is element-wise multiplication, $*$ is convolution, and $U, W,$ $U_r, W_r, U_z, W_z$ are the 2D convolutional kernels, whose parameters are optimized end-to-end.  Gate $r_t$ masks elements of $M_{t-1}$ (i.e., it allows the previous state to be forgotten) to generate candidate memory $\tilde{M}_t$. And gate $z_t$ determines how to weight and combine the memory from the previous step $M_{t-1}$ with the candidate memory $\tilde{M}_t$, to generate the new memory $M_t$.

\begin{wrapfigure}{r}{0.42\textwidth}
  \centering
  \includegraphics[width=0.38\textwidth]{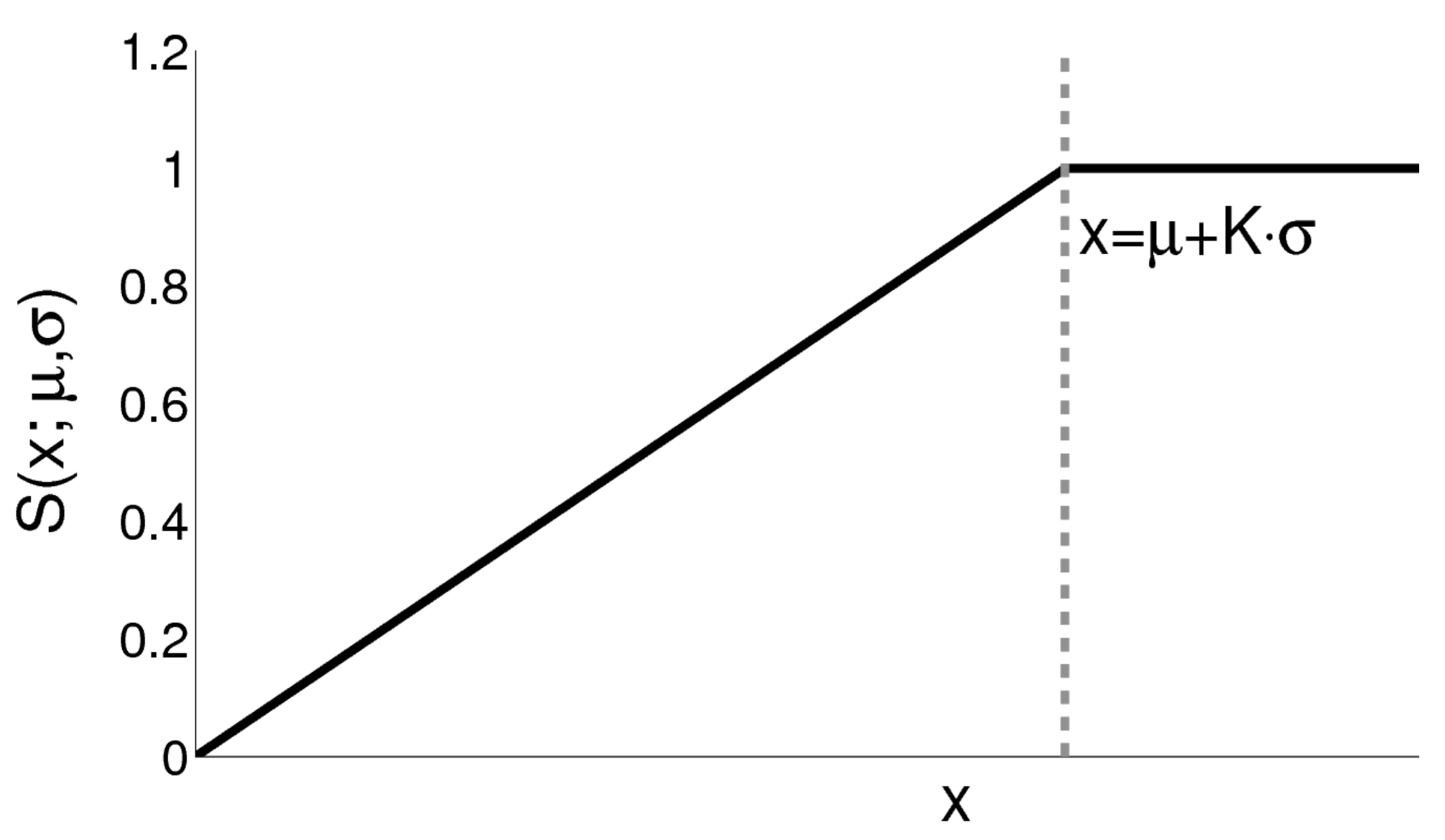}
  \caption{$S(x; \mu,\sigma)$ squashes any value in $[0, +\inf)$ into range $[0, 1]$, with a linear scaling function thresholded at $\mu + K \cdot \sigma$. We set $K=3$.}
  \label{fig:AdaScale}
\end{wrapfigure}
To generate $r_t$ and $z_t$, the STMM first computes an affine transformation of $M_{t-1}$ and $F_t$, and then ReLU~\cite{krizhevsky-nips2012} is applied to the outputs.  Since $r_t$ and $z_t$ are gates, their values need to be in the range of $[0, 1]$. Therefore, we make two changes to the standard BatchNorm~\cite{ioffe-batchnorm2015} (and denote it as ${\tt BN^*}$) such that it normalizes its input to $[0, 1]$, instead of zero mean and unit standard deviation.

First, our variant of BatchNorm computes the mean $\mu(X)$ and standard deviation $\sigma(X)$ for an input batch $X$, and then normalizes values in $X$ with the linear squashing function $S(X; \mu, \sigma)$ shown in Fig.~\ref{fig:AdaScale}. Second, we compute the mean and standard deviation for each batch independently instead of keeping running averages across training batches.  In this way, we do not need to store different statistics for different time-steps, which allows us to generate test results for sequence lengths not seen during training (e.g., we can compute detections on longer sequences than those seen during training as demonstrated in Sec.~\ref{sect:dissect}). Note that a key difference between ${\tt BN^*}$ and instance/layer normalization~\cite{huang-iccv2017,ba-layernorm2016} is that ${\tt BN^*}$ guarantees that \emph{each and every} value in its output is normalized within $[0, 1]$ (which is necessary for gating variables), whereas neither instance nor layer normalization ensures this property. Although simple, we find ${\tt BN^*}$ works well for our purpose.

\subsubsection{Differences with ConvGRU~\cite{ballas-iclr2016}} A key practical challenge of learning video object detectors is the lack of intra-category object diversity in contemporary video datasets; i.e., since video frames are highly redundant, a video dataset of e.g., 1 million frames has much lower diversity than an image dataset with 1 million images.  The cost of annotation is much higher in video, which makes it difficult to have the same level of diversity as an image dataset.  Therefore, transferring useful information from large-scale \emph{image} datasets like ImageNet~\cite{imagenet}---into the memory processing unit itself---would benefit our video object detector by providing additional diversity.

Specifically, we would like to initialize our STMN detector with the weights of the state-of-the-art static image-based RFCN detector~\cite{dai-nips2016} which has been pretrained on ImageNet DET images, and continue to fine-tune it on ImageNet VID videos.  In practice, this would entail converting the last convolution layer before the Position-Sensitive ROI pooling in RFCN into our STMM memory unit (see Fig.~\ref{fig:architecture}).  However, this conversion is non-trivial with standard recurrent units like LSTM/GRU that employ {\tt Sigmoid}/{\tt Tanh} nonlinearities, since they are different from the {\tt ReLU} nonlinearity employed in the R-FCN convolutional layers.

Thus, to transfer the weights of the pre-trained RFCN static image detector into our STMN video object detector, we make two changes to the ConvGRU~\cite{ballas-iclr2016}.  First, in order to make full use of the pre-trained weights, we need to make sure the output of the recurrent computation unit is compatible with the pre-trained weights before and after it. As an illustrative example, since the output of the standard ConvGRU is in $[-1, 1]$ (due to {\tt Tanh} non-linearity), there would be a mismatch with the input range that is expected by the ensuing pre-trained convolutional layer (the expected values should all be positive due to {\tt ReLU}). To solve this incompatibility, we change the non-linearities in standard ConvGRU from {\tt Sigmoid} and {\tt Tanh} to {\tt ReLU}.  Second, we initialize $W_z$, $W_r$ and $W$ in Eqs.~\ref{eq:SMMz}-\ref{eq:SMMCand} with the weights of the convolution layer that is swapped out, rather than initializing them with random weights. Conceptually, this can be thought of as a way to initialize the memory with the pre-trained static convolutional feature maps.  In Sec.~\ref{sect:dissect}, we show that these modifications allow us to make better use of pre-trained weights and achieve better detection performance.

\subsection{Spatial-temporal memory alignment}  
\begin{figure*}[t!]
    \centering
    \includegraphics[width=1\textwidth]{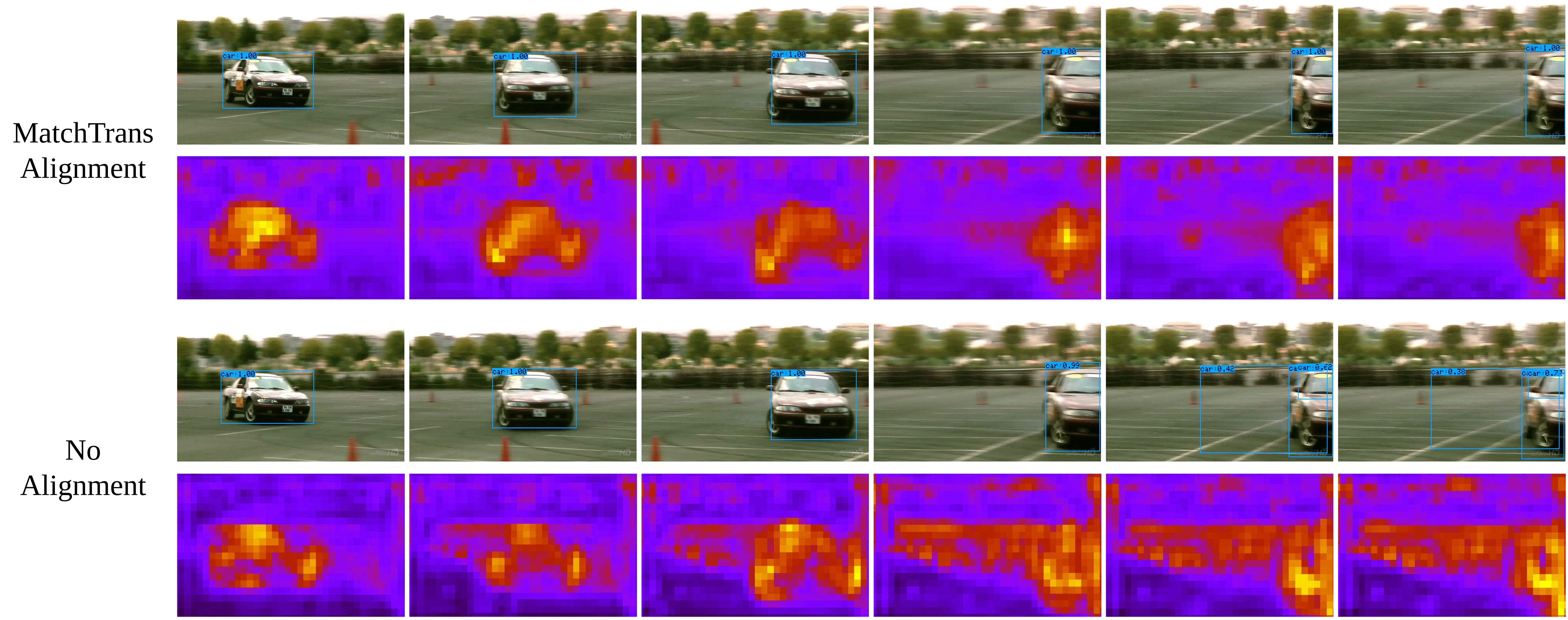}
    \caption{Effect of alignment on spatial-temporal memory. In the first and second rows, we show the detection and the visualization of the spatial-temporal memory (by computing the $L2$ norm across feature channels at each spatial location to get a saliency map), respectively, with MatchTrans alignment. The detection and memory without alignment are shown in rows 3 and 4, respectively.  Without proper alignment, the memory has a hard time forgetting an object after it has moved to a different spatial position (third row), which is manifested by a trail of saliency on the memory map due to overlaying multiple unaligned maps (fourth row).  Alignment with MatchTrans helps generate a much cleaner memory (second row), which also results in better detections (first row). Best viewed in pdf.}
    \label{fig:alignment}
\end{figure*}

Next, we explain how to align the memory across frames.  Since objects \emph{move} in videos, their spatial features can be mis-aligned across frames. For example, the position of a bicycle in frame $t-1$ might not be aligned to the position of the bicycle in frame $t$ (as in Fig.~\ref{fig:architecture}). In our case, this means that the spatial-temporal memory $M_{t-1}$ may not be spatially aligned to the feature map for current frame $F_t$. This can be problematic, for example in the case of Fig.~\ref{fig:alignment}; without proper alignment, the spatial-temporal memory can have a hard time forgetting an object after it has moved to a different spatial position. This is manifested by a trail of saliency, in the fourth row of Fig.~\ref{fig:alignment}, due to the effect of overlaying multiple unaligned feature maps.  Such hallucinated features can lead to false positive detections and inaccurate localizations, as shown in the third row of Fig.~\ref{fig:alignment}.

\begin{figure*}[t!]	
	\centering
	\includegraphics[width=0.95\textwidth]{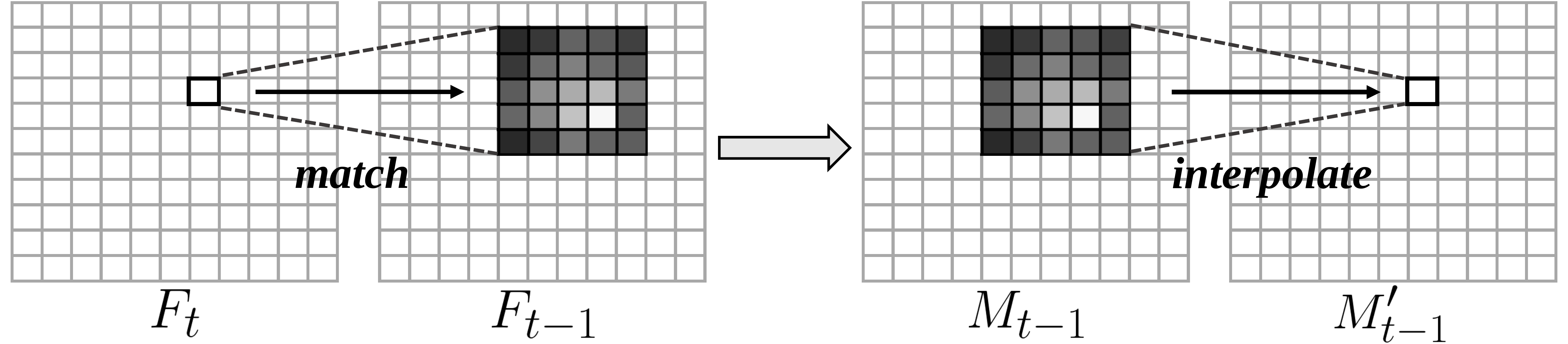}
	\caption{The transformation coefficients $\Gamma$ for position $(x,y)$ are computed by matching $F_t(x,y)$ to $F_{t-1}(i,j)$, where $i,j$ indexes a spatial neighborhood surrounding $(x,y)$.  The transformation coefficients are then used to synthesize $M_{t-1}'(x,y)$ by interpolating the corresponding $M_{t-1}(i,j)$ feature vectors.}
	\label{fig:matchTrans}
\end{figure*}

To alleviate this problem, we propose the MatchTrans module to align the spatial-temporal memory across frames.  For a feature cell $F_t(x,y) \in 1 \times 1 \times D$ at location $(x,y)$ in $F_t$, MatchTrans computes the affinity between $F_t(x,y)$ and feature cells in a small vicinity around location $(x,y)$ in $F_{t-1}$, in order to transform the spatial-temporal memory $M_{t-1}$ to align with frame $t$.  More formally, the transformation coefficients $\Gamma$ are computed as:
\begin{align*}
\Gamma_{x,y}(i,j) = \frac{F_t(x,y) \cdot F_{t-1}(x+i,y+j)}{\sum_{i,j \in \{-k,\dots, k\}}F_t(x,y) \cdot F_{t-1}(x+i,y+j)},
\end{align*}
where both $i$ and $j$ are in the range of $[-k, k]$, which controls the size of the matching vicinity. With $\Gamma$, we transform the unaligned memory $M_{t-1}$ to the aligned $M_{t-1}'$ as follows:
\begin{align*}
M_{t-1}'(x,y) = \sum_{i,j \in \{-k,\dots,k\}} \Gamma_{x,y}(i,j) \cdot M_{t-1}(x+i,y+j).
\end{align*}
The intuition here is that given transformation $\Gamma$, we reconstruct the spatial memory $M_{t-1}'(x,y)$ as a weighted average of the spatial memory cells that are within the $(2k+1) \times (2k+1)$ vicinity around $(x,y)$ on $M_{t-1}$; see Fig.~\ref{fig:matchTrans}. At this point, we can thus simply replace all occurrences of $M_{t-1}$ with the spatially aligned memory $M_{t-1}'$ in Eqs.~\ref{eq:SMMz}-\ref{eq:SMMFinalM}. With proper alignment, our generated memory is much cleaner (second row of Fig.~\ref{fig:alignment}) and leads to more accurate detections (first row of Fig.~\ref{fig:alignment}).  Since the computational cost is quadratic in $k$, we set $k=2$ for all our experiments as this choice provides a good trade-off between performance and computation.

Our MatchTrans is related to the alignment module used in recent video object detection work by~\cite{zhu-iccv2017,zhu-cvpr2018}. However,~\cite{zhu-iccv2017,zhu-cvpr2018} use optical flow, which needs to be computed either externally e.g., using~\cite{Brox-pami2011}, or in-network through another large CNN e.g., FlowNet~\cite{flownet}.  In contrast, our MatchTrans is much more efficient, saving computation time and/or space for storing optical flow. For example, it is nearly an order of magnitude faster to compute (on average, 2.9ms vs.~24.3ms for an 337x600 frame) than FlowNet~\cite{flownet}, which is one of the fastest optical flow methods.  Also, a similar procedure for computing transformation coefficients was used in~\cite{feichtenhofer-iccv2017}. However, in~\cite{feichtenhofer-iccv2017}, the coefficients are fed as input to another network to regress the displacement of bounding boxes for tracking, whereas we use it to warp the entire feature map for aligning the memory.  In other words, rather than use the transformation coefficients to track and connect detections, we instead use them to align the memory over time to produce better features for each candidate object region.  We show in Sec.~\ref{sect:sota} that this leads to better performance on ImageNet VID.

\subsection{Temporal linkage during testing} Finally, even though we enforce temporal smoothness in our spatial-temporal memory (i.e., at the feature level), we do not have an explicit smoothness constraint in the output space to ensure that detections in adjacent frames are spatially smooth. We therefore apply standard Seq-NMS~\cite{han-arxiv2016} over our per-frame detections, following~\cite{zhu-iccv2017,feichtenhofer-iccv2017}.

\subsection{Approach summary}  Through the specially designed Spatial-Temporal Memory and MatchTrans modules, our STMN detector aggregates and aligns useful information from temporally nearby frames for video object detection.

\section{Results}\label{sect:results}

We show quantitative and qualitative results of our STMN video object detector, and compare to both state-of-the-art static image and video detectors. We also conduct ablation studies to analyze the different components of our model.

\paragraph{Dataset.} We use ImageNet VID~\cite{imagenetvid}, which has 3862/555/937 videos for training/validation/testing for 30 categories. Bounding box annotation is provided for all frames. We choose ImageNet VID for its relatively large size as well as for ease of comparison to existing state-of-the-art methods~\cite{imagenetvid,dai-nips2016,kang-tcnn2017,kang-cvpr2017,feichtenhofer-iccv2017,zhu-iccv2017,zhu-cvpr2018}.

\paragraph{Implementation details.} For object proposals, we use DeepMask~\cite{pinheiro-nips2015} for our method and our own baselines. We use the R-FCN detector~\cite{dai-nips2016} with ResNet-101~\cite{he-cvpr2016} as the backbone network. Following~\cite{feichtenhofer-iccv2017}, we first train R-FCN on ImageNet DET, and then transfer its weights (using the method described in Sec.~\ref{sect:stmm}) to initialize our STMN detector and continue fine-tuning it on ImageNet VID. We set sequence length $T=7$ during training. For testing, we observe better performance when using a longer sequence length; specifically, $T=11$ frames provides a good balance between performance and GPU memory/computation (we later show the relationship between performance and test sequence length). We set the number of channels of the spatial memory to 512. To reduce redundancy within sequences, we form a sequence by sampling 1 in every 10 video frames with uniform stride. For training, we start with a learning rate of 1e-3 with SGD and lower it to 1e-4 when training loss plateaus. During testing we ensemble the detection results of the STMN detector with the initial R-FCN detector from which it started since it comes for free as a byproduct of the training procedure. We employ standard left-right flipping augmentation.

\begin{table}[t!]
\centering
\tabcolsep=0.1cm
    \begin{tabular}{l  c c c c }
        & Base network
        & Base detector
        & \shortstack{Test}
        & \shortstack{Val}
        \\
        \hline
        STMN (Ours) & ResNet-101  & R-FCN &  -  & \textbf{80.5}
        \\
        D\&T~\cite{feichtenhofer-iccv2017} & ResNet-101 & R-FCN & - & 79.8
        \\
        Zhu et al.~\cite{zhu-cvpr2018} & ResNet-101+DCN & R-FCN & - & 78.6
        \\
        FGFA~\cite{zhu-iccv2017} & ResNet-101 & R-FCN & - & 78.4
        \\
        T-CNN~\cite{kang-tcnn2017} & DeepID+Craft~\cite{ouyang-arxiv2014,yang-cvpr2016} & RCNN & 67.8 & 73.8
        \\
        R-FCN~\cite{dai-nips2016} & ResNet-101 & R-FCN & - & 73.4
        \\
        TPN~\cite{kang-cvpr2017} & GoogLeNet & TPN & - & 68.4
        \\
        \hline
        STMN (Ours) & VGG-16  & Fast-RCNN &  \textbf{56.5}  & \textbf{61.7}
        \\
        Faster-RCNN~\cite{imagenetvid,han-arxiv2016} & VGG-16  & Faster-RCNN & 48.2 & 52.2
        \\
        ITLab VID - Inha~\cite{imagenetvid} & VGG-16 & Fast-RCNN & 51.5 & -
        \\
        \hline

    \end{tabular}
    \caption{mAP comparison to the state-of-the-art on ImageNet VID. For both the ``R-FCN+ResNet-101" and the ``Fast-RCNN+VGG-16" settings, our STMN detector outperforms all existing methods with the same base detector and backbone network.  Furthermore, in both cases, our STMN outperforms the corresponding static-image detector by a large margin.}
    \label{table:externalquant}
\end{table}

\subsection{Comparison to state-of-the-art}\label{sect:sota}

Table~\ref{table:externalquant} shows the comparison to existing state-of-the-art image and video detectors. First, our STMN detector outperforms the static-image based R-FCN detector with a large margin (+7.1\%). This demonstrates the effectiveness of our proposed spatial-temporal memory. Our STMN detector also achieves the best performance compared to all existing video object detection methods with ResNet-101 as the base network.  Furthermore, in order to enable a fairer comparison to older methods that use Fast/Faster-RCNN + VGG-16 as the base detector and backbone network, we also train an STMN model with the Fast-RCNN as the base detector and VGG-16 as the backbone feature network. Specifically, we first train a static-image Fast-RCNN detector and initialize the weights of STMN using a similar procedure as described in Sec.~\ref{sect:stmm}.\footnotemark~With this setting, our STMN achieves 61.7\% val mAP, which is much higher than its static-image based counterpart (52.2\%).  This result shows that our method can be generalized across different base detectors and backbone networks.

\footnotetext{Specifically, we convert the {\tt conv5} layer in VGG-16 to an STMM module by initializing $W_z$, $W_r$ and $W$ in Eqs.~\ref{eq:SMMz}-\ref{eq:SMMCand} with the weights of {\tt conv5}.}

When examining per-category results, our method shows the largest improvement on categories like ``sheep'', ``rabbit'', and ``domestic cat'' compared to methods like~\cite{feichtenhofer-iccv2017}. In these cases, we see a clear advantage of aggregating information across multiple frames (vs.~2 frames as in~\cite{feichtenhofer-iccv2017}), as there can be consecutive ``hard" frames spanning multiple ($>2$) frames (e.g., a cat turning away from the camera for several frames). On the other hand, we find that the three categories on which we perform the worst are ``monkey", ``snake", and ``squirrel".  These are categories with large deformation and strong motion blur.  When the per-frame appearance features fail to accurately model these objects due to such challenges, aggregating those features over time with our STMM does not help.  Still, overall, we find that our model produces robust detection results across a wide range of challenges as demonstrated next in the qualitative results.

\begin{figure*}[tp!]
    \centering
    \includegraphics[width=0.97\textwidth]{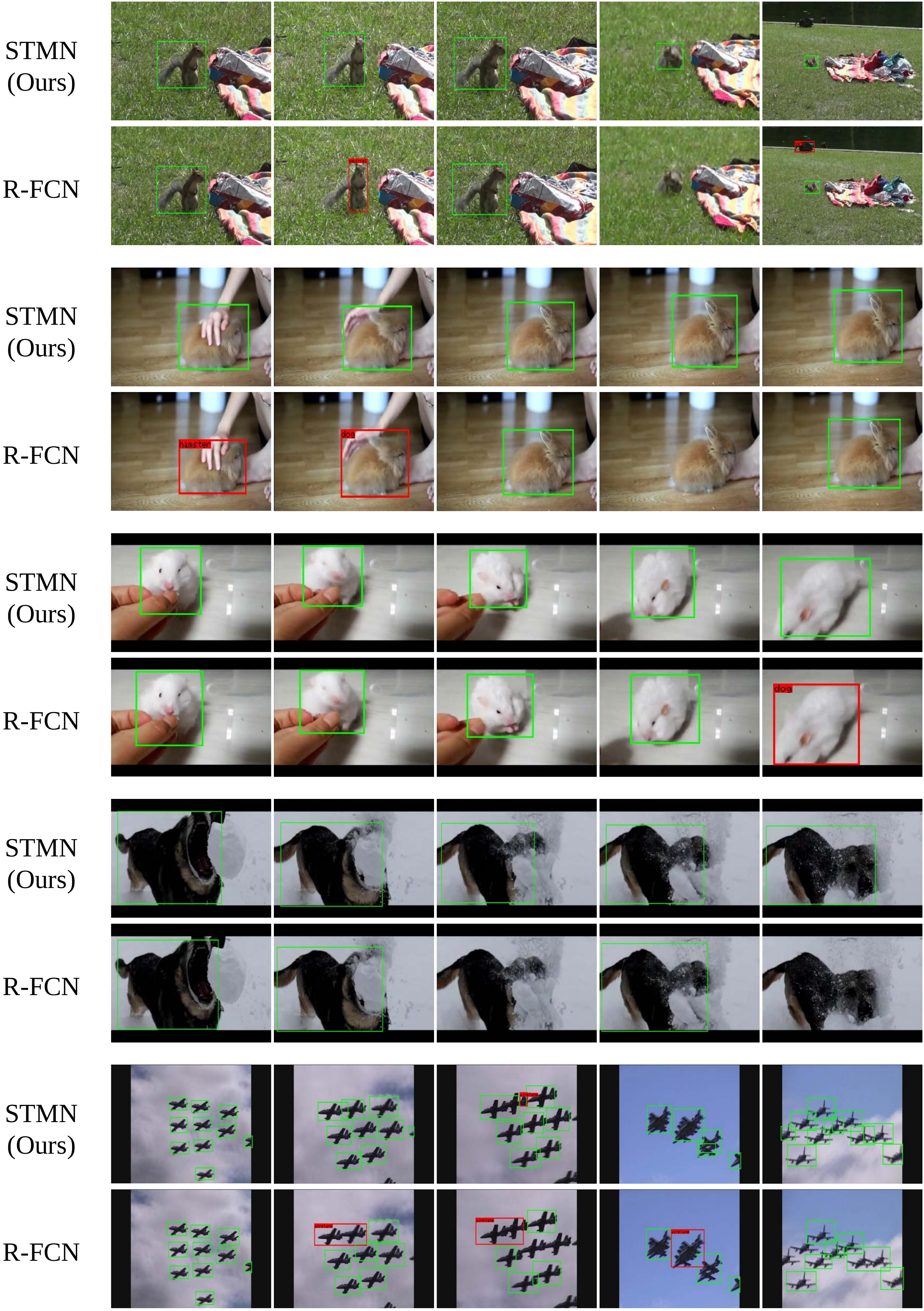}
    \caption{Example detections produced by our STMN video object detector vs.~R-FCN image detector. Green and red boxes indicate correct and incorrect detections, respectively.  For any false positive detection due to misclassification or mislocalization, the predicted category label is shown at the top-left corner of the box.  The ground-truth object in each sequence is: ``squirrel'', ``rabbit'', ``hamster'', ``dog,'' and ``airplane".  Best viewed in pdf, zoomed-in.}  
    \label{fig:detection}
\end{figure*}

\subsection{Qualitative results} Fig.~\ref{fig:detection} shows qualitative comparisons between our STMN detections and the static image R-FCN detections.  Our STMN detections are more robust to motion blur; e.g., in the last frame of the ``hamster'' sequence, R-FCN gets confused about the class label of the object due to large motion blur, whereas our STMN detector correctly detects the object.  In the case of difficult viewpoint and occlusion (``dog'' and ``rabbit'', respectively), our STMN produces robust detections by leveraging the information from neighboring easier frames (i.e., center frame in the ``rabbit" sequence and the first frame in the ``dog" sequence). Also, our model outputs detections that are more consistent across frames, compared with the static image detector, as can be seen in the case of ``squirrel'' and ``rabbit".   Finally, our STMN detector is also more robust in crowded scenes as shown in the ``airplane" sequence.

\subsection{Ablation studies}\label{sect:dissect} We next conduct ablation studies to analyze the impact of each component in our model by comparing it to a number of baselines that lack one or more components. For this, we use Fast-RCNN as the base detector and VGG-16 as the backbone network since it is much faster to train compared to RFCN + ResNet-101. To ensure a clean analysis, we purposely do not employ any data augmentation during training for this ablative study. 

\subsubsection{Contribution of STMN components}

The first baseline, compared with our model, lacks the MatchTrans module and thus does not align the memory from frame to frame (STMN-No-MatchTrans). The second baseline computes the memory using ConvGRU~\cite{ballas-iclr2016}, instead of our proposed STMM. Like ours, this baseline (ConvGRU-Pretrain) also uses pre-trained ImageNet weights for both the feature stack and prediction layers.  Our final baseline is ConvGRU without pre-trained weights for the ensuing prediction FCs (ConvGRU-FreshFC).

\begin{table*}[t!]
\centering
    {
    \begin{tabular}{l | c | c | c | c}
        \hline
        & {STMN}
        &  {\makecell{STMN\\No-MatchTrans}}
        &  {\makecell{ConvGRU\\Pretrain}}
        &  {\makecell{ConvGRU\\FreshFC}}
        \\
        \hline
        Test mAP & 50.7 & 49.0 & 48.0 & 44.8
        \\
        \hline
    \end{tabular}
    }
    \caption{Ablation studies on ImageNet VID. Our improvements over the baselines show the importance of memory alignment across frames with MatchTrans (vs. STMN-No-MatchTrans), and the effectiveness of using pre-trained weights with STMM over standard ConvGRU (vs. ConvGRU-Pretrain and ConvGRU-FreshFC).}
    \label{table:quant}
\end{table*}

Table~\ref{table:quant} shows the results.  First, comparing our STMN to the STMN-No-MatchTrans baseline, we observe a 1.7\% test mAP improvement brought by the spatial alignment across frames.  This result shows the value of our MatchTrans module.  To compare our STMM with ConvGRU, we first replace STMM with ConvGRU and as with standard practice, randomly initialize the weights for the FC layers after the ConvGRU.  With this setting (ConvGRU-FreshFC), we obtain a relatively low test mAP of 44.8\%, due to the lack of data to train the large amount of weights in the FCs. This result shows that initializing the memory by only partially transferring the pre-trained ImageNet weights is suboptimal. If we instead initialize the weights of the FCs after the ConvGRU with pre-trained weights (ConvGRU-Pretrain), we improve the test mAP from 44.8\% to 48.0\%.  Finally, by replacing {\tt Sigmoid} and {\tt Tanh} with {\tt ReLU}, which is our full model (STMN), we boost the performance even further to 50.7\%. This shows the importance of utilizing pre-trained weights in both the feature stacks and prediction head, and the necessity of an appropriate form of recurrent computation that best matches its output to the input expected by the pre-trained weights.

\subsubsection{Length of test window size}\label{sect:windowlength} We next analyze the relationship between detection performance and length of test window size.  Specifically, we test our model's performance with test window size 3, 7, 11, and 15, on ImageNet VID validation set (the training window size is always 7). The corresponding mAP differences, with respect to that of window size 7, are -1.9\%, 0.0\%, +0.7\%, +1.0\%, respectively; as we increase the window size, the performance tends to keep increasing.  This suggests the effectiveness of our memory: the longer the sequence, the more longer-range useful information is stored in the memory, which leads to better detection performance.  However, increasing the test window size also increases computation cost and GPU memory consumption. Therefore, we find that setting the test window size to 11 provides a good balance.

\subsection{Computational overhead of STMN} Finally, we sketch the computational overhead of our memory module. To forward a batch of 11 frames of size 337x600, it takes 0.52 and 0.83 seconds for R-FCN and STMN respectively, on a Titan X GPU. The added 0.028 (=0.31/11) secs/frame is spent on STMM computation including MatchTrans alignment.

\section{Conclusion}

We proposed a novel spatial-temporal memory network (STMN) for video object detection.  Our main contributions are a carefully-designed recurrent computation unit that integrates pre-trained image classification weights into the memory and an in-network alignment module that spatially-aligns the memory across time.  Together, these lead to state-of-the-art results on ImageNet VID.  Finally, we believe that our STMN could also be useful for other video understanding tasks that require accurate spatial information like action detection and keypoint detection.

\paragraph{Acknowledgments} This work was supported in part by the ARO YIP W911NF17-1-0410, NSF CAREER IIS-1751206, AWS Cloud Credits for Research Program, and GPUs donated by NVIDIA. The views and conclusions contained in this document are those of the authors and should not be interpreted as representing the official policies, either expressed or implied, of ARO or the U.S. Government. The U.S. Government is authorized to reproduce and distribute reprints for Government purposes notwithstanding any copyright notation herein.

%
%
%
%
%
%
%

\bibliographystyle{splncs04}
\bibliography{strings,bibs}

\end{document}